\documentclass[
]{ceurart}

\usepackage{listings}
\usepackage{booktabs}

\usepackage{graphicx}
\usepackage{subcaption}

\begin{document}

\copyrightyear{2026}
\copyrightclause{Copyright for this paper by its authors.
  Use permitted under Creative Commons License Attribution 4.0
  International (CC BY 4.0).}

\conference{Twelfth Italian Conference on Computational Linguistics (CLiC-it 2026),
  September 14--16, 2026, Palermo, Italy}

\title{Larry Caused the Car to Stop, But the Model Didn't Notice: Transformer Blindness to the M-Heuristic}

\author[1]{\c{S}tefania Butnaru}[%
email=stefania.butnaru@s.unibuc.ro,
]
\author[2,3,4]{Claudiu Creanga}[%
email=claudiu.creanga@fmi.unibuc.ro,
]
\cormark[1]
\author[2,4]{Liviu P. Dinu}[%
email=ldinu@fmi.unibuc.ro,
]

\address[1]{Faculty of Foreign Languages, University of Bucharest, Romania}
\address[2]{Faculty of Mathematics and Computer Science, University of Bucharest, Romania}
\address[3]{Interdisciplinary School of Doctoral Studies, University of Bucharest, Romania}
\address[4]{HLT Research Center, University of Bucharest, Romania}

\cortext[1]{Corresponding author.}

\begin{abstract}
  Modern transformer models excel at capturing semantic relationships through sentence embeddings, yet their ability to perform pragmatic reasoning remains understudied. This paper investigates whether encoder-based transformers such as DeBERTa employ the M-Heuristic (the neo-Gricean principle that marked linguistic forms implicate marked meanings). We test this hypothesis by contrasting lexical causatives (e.g., ``Larry stopped the car'') with periphrastic causatives (e.g., ``Larry caused the car to stop'') using a Natural Language Inference framework. Our experiments across 188 conditions with 15 ambitransitive verbs reveal that DeBERTa, RoBERTa, and BART show no evidence of capturing the pragmatic distinction between these forms, with DeBERTa predicting ``Neutral'' for 100\% of cases. Probing analysis initially suggested a representation-use dissociation, but control experiments reveal the probe was tracking syntactic complexity, not causative pragmatics. Semantic similarity over 30 triplets places periphrastic causatives closer to \textit{unmediated} manner descriptions in 29/30 cases, opposite to M-Heuristic predictions in the embedding space. Under explicit metalinguistic framing, Gemini Flash-Lite reaches 100\% with item-specific traces, so the principle is available under instruction yet unused in default NLI.
\end{abstract}

\begin{keywords}
  pragmatics \sep
  M-Heuristic \sep
  Natural Language Inference \sep
  causatives \sep
  transformers \sep
  DeBERTa
\end{keywords}

\maketitle

\section{Introduction}

Human communication depends on recovering speaker intentions beyond what is explicitly stated. While articulation is costly, inference is relatively cheap, leading speakers to rely on implicit meaning and listeners to employ systematic heuristics to narrow down communicative intent \cite{Levinson2000}. This pragmatic dimension of language understanding (deriving meaning from \textit{how} something is said, not only \textit{what} is said) remains difficult for computational systems.

Modern transformer architectures, particularly encoder-only models like DeBERTa \cite{he2021deberta} with disentangled attention mechanisms, have strong performance on semantic understanding tasks. These models excel at sentence embeddings where relative position matters and anchor many Natural Language Inference (NLI) systems. Their capacity for pragmatic reasoning (inferences from linguistic form rather than content alone) remains largely unexplored.

This paper investigates whether state-of-the-art sentence embedders employ what Levinson terms the \textbf{M-Heuristic}: the principle that ``what is said in an abnormal way isn't normal'' \cite{Levinson2000}. According to this neo-Gricean framework, marked (unusual, complex) linguistic forms implicate marked (unusual, non-stereotypical) meanings, while unmarked forms pick up standard interpretations. The M-Heuristic is triggered by form itself, unlike scalar implicatures (Q-Heuristics) that depend on lexical scales.

To test this hypothesis, we focus on the contrast between \textbf{lexical causatives} and \textbf{periphrastic causatives} (a domain where form-meaning correspondences are well-documented in linguistic theory and human experimental data). Consider the following pair:

\begin{figure}[t]
  \centering
  \includegraphics[width=\linewidth]{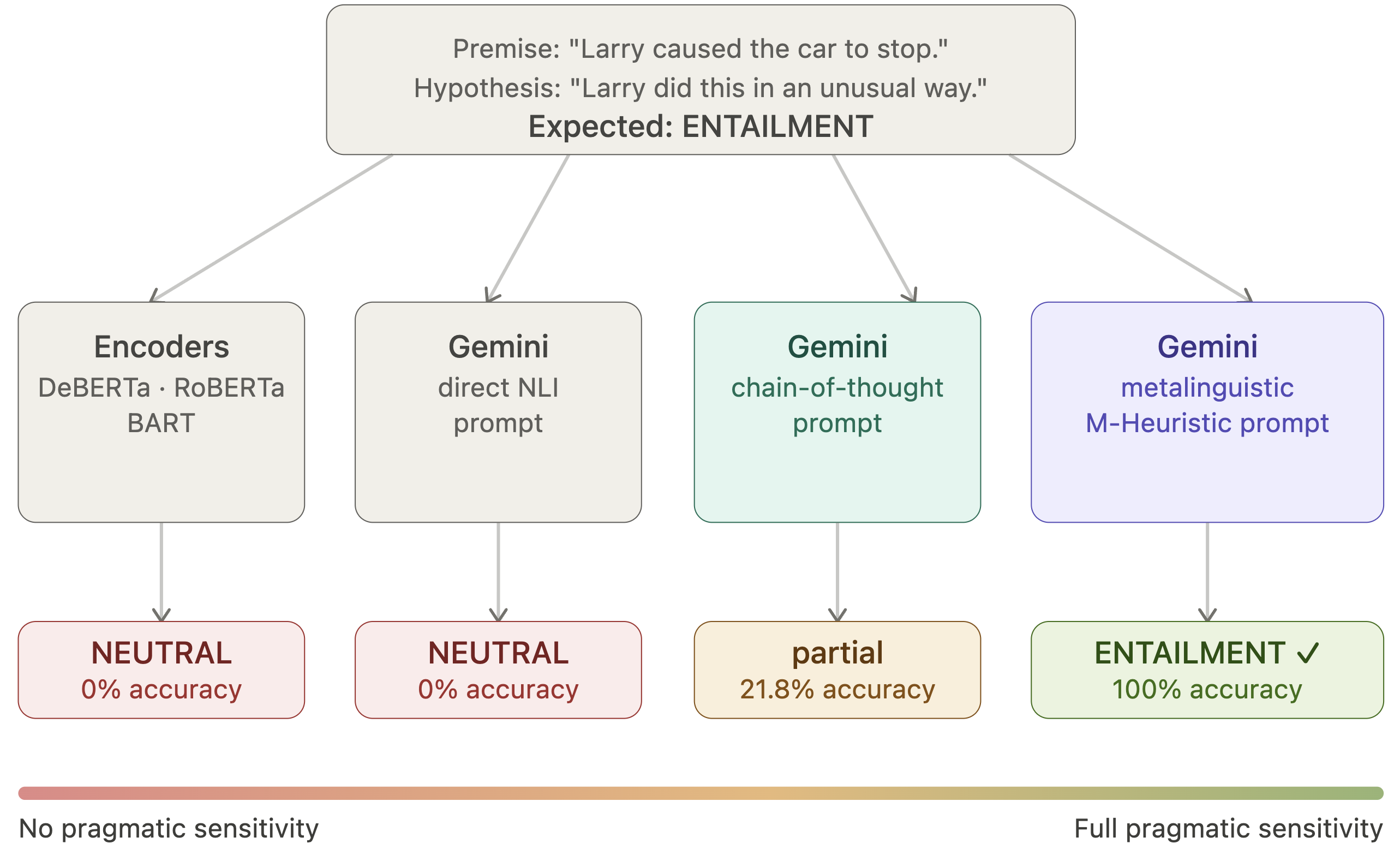}
  \caption{One fixed NLI instance from our tests (M-Heuristic gold: \textbf{Entailment}). DeBERTa, RoBERTa, BART, and Gemini under direct NLI all predict \textbf{Neutral} on this item. Across all 188 items: chain-of-thought recovers 21.8\%; metalinguistic prompting reaches 100\%.}
  \label{fig:teaser}
\end{figure}

\begin{enumerate}
    \item[(1)] \textit{Larry stopped the car.} (lexical causative)
    \item[(2)] \textit{Larry caused the car to stop.} (periphrastic causative)
\end{enumerate}

\noindent While both sentences describe causation, they carry distinct pragmatic implications. The unmarked lexical form (1) I-implicates direct, stereotypical causation (e.g., by pressing the brake pedal), whereas the marked periphrastic form (2) M-implicates indirect or unusual causation (e.g., by jumping in front of the vehicle). Human experimental data from \citeauthor{Wolff2003} confirms this intuition: participants consistently select lexical descriptions for unmediated causal events and periphrastic descriptions for mediated ones.

We operationalize this test using Natural Language Inference, an empirical framework now widely employed for probing pragmatic competence in language models \cite{Jeretic2020,Ma2025}. Our experiments across 188 conditions (spanning 15 ambitransitive verbs and 3 contexts) show that \textbf{none of the tested transformer models replicate human M-Heuristic expectations}. DeBERTa predicted ``Neutral'' for 100\% of premise-hypothesis pairs; RoBERTa and BART were Neutral-heavy and missed pragmatic entailments.

These findings expose a limitation in current NLI models: they excel at semantic processing but not at the form-meaning correspondences that underlie manner implicatures. Unlike scalar implicatures, which can be learned from distributional patterns involving lexical triggers like ``some'' and ``all,'' M-implicatures require reasoning about \textit{why} a speaker chose a marked form when an unmarked alternative was available, a level of pragmatic sophistication that current architectures do not exhibit.

\subsection{Contributions}

We contribute:
\begin{itemize}
  \item the \textbf{first NLI-based evaluation} of the M-Heuristic in transformer models, complementing prior surprisal-based work \cite{Cong2024};
  \item a \textbf{controlled framework} using lexical and periphrastic causatives across 15 ambitransitive verbs, grounded in linguistic theory \cite{Levinson2000} and human behavioral data \cite{Wolff2003};
  \item evidence that \textbf{state-of-the-art NLI models show no sensitivity} to the pragmatic distinction encoded in causative form;
  \item probing and semantic similarity analysis demonstrating that DeBERTa encodes \textbf{syntactic form} but not \textbf{pragmatic meaning};
  \item a \textbf{prompt comparison} showing Gemini Flash-Lite progresses from 0\% under direct NLI to 100\% under metalinguistic framing, consistent with pragmatic knowledge present in model weights but absent from default processing (Figure~\ref{fig:teaser}).
\end{itemize}
We will release the code and data upon acceptance at: \url{https://github.com/ClaudiuCreanga/m-heuristic}.

\section{Related Work}

Our theoretical foundation lies in Levinson's neo-Gricean framework \cite{Levinson2000}, which systematizes Grice's maxims into three heuristics. The \textbf{Q-Heuristic} governs scalar implicatures (``some'' implicates ``not all''). The \textbf{I-Heuristic} dictates that unmarked expressions warrant stereotypical interpretations. The \textbf{M-Heuristic}, central to our investigation, derives from Grice's Maxim of Manner: what is said in an abnormal way implicates an abnormal situation. Unmarked forms trigger I-inferences; marked forms trigger M-inferences.

Causation provides an ideal testing ground for the M-Heuristic. Following \citet{Wolff2003}, \textbf{lexical causatives} (e.g., ``Larry stopped the car'') are syntactically compact and map to unmediated causal chains, I-implicating direct causation. \textbf{Periphrastic causatives} (``Larry caused the car to stop'') are structurally complex and map to mediated chains, M-implicating indirect causation. Human experiments confirm this mapping \cite{Wolff2003}.

NLI has become the standard framework for evaluating pragmatic competence \cite{Bowman2015}. As the scope of NLI expands into highly specialized domains, from clinical reasoning \cite{micluta-campeanu-etal-2024-unibuc} to pragmatic inference, researchers are increasingly exploring how targeted prompting and pre-trained models handle these nuanced tasks. Yet, standard corpora contain almost no pragmatic inference examples \cite{Jeretic2020}. The IMPPRES dataset \cite{Jeretic2020} showed BERT can learn some scalar implicatures, though often via statistical heuristics rather than pragmatic reasoning \cite{Yue2024}.
\citet{Williams2020} provide a fine-grained hand-annotation of Adversarial NLI development data by inference type, supporting more targeted analysis of model behaviour on NLI benchmarks.

Unlike Q-implicatures triggered by specific lexical items, M-implicatures are triggered by \textit{structural markedness} (a comparison between the form used and the simpler alternative). This makes them harder for distributional models. \citet{Cong2024} found pre-trained models performed at or below chance on manner implicatures using surprisal scores. RSA frameworks \cite{Bergen2016} can model M-implicatures but require explicit utterance costs. We extend that line with an NLI study of causative alternations.

\section{Experimental Setup}

We first summarize the human baseline (\citet{Wolff2003}): listeners link causative wording to mediated vs.\ unmediated causal chains. We then give the NLI design that adapts that paradigm.

\subsection{Human Baseline: Wolff's Causative Language Experiments}

\citet{Wolff2003} motivated the contrast with behavioral experiments on causative language. His \textbf{no-intervening-cause hypothesis} distinguishes \textbf{unmediated} causal chains (the causer acts directly on the patient) from \textbf{mediated} chains (those in which an intervening variable (another agent, mechanism, or temporal gap) separates cause from effect).

\begin{figure}[ht]
  \centering
  \begin{subfigure}[t]{0.48\linewidth}
    \centering
    \includegraphics[width=\linewidth]{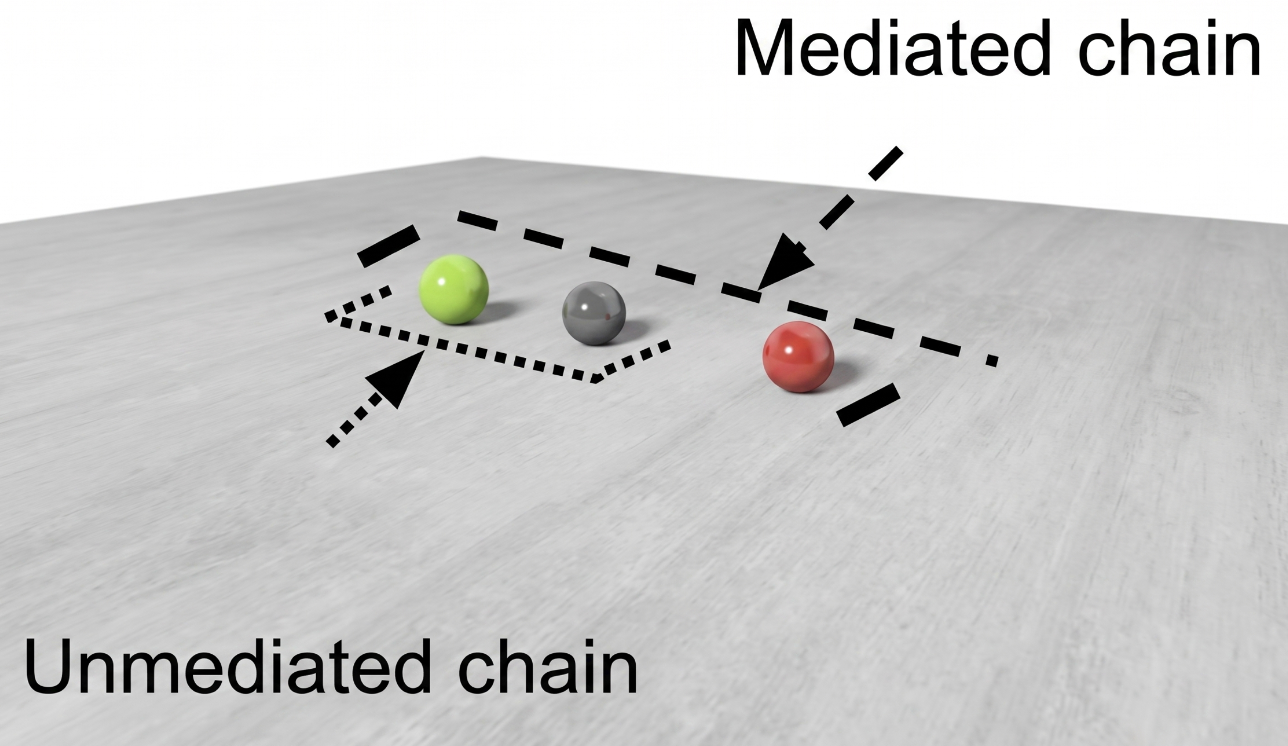}
    \caption{Mediated vs.\ unmediated causal chains from \citet{Wolff2003}.}
    \label{fig:causal-chains}
  \end{subfigure}\hfill
  \begin{subfigure}[t]{0.48\linewidth}
    \centering
    \includegraphics[width=\linewidth]{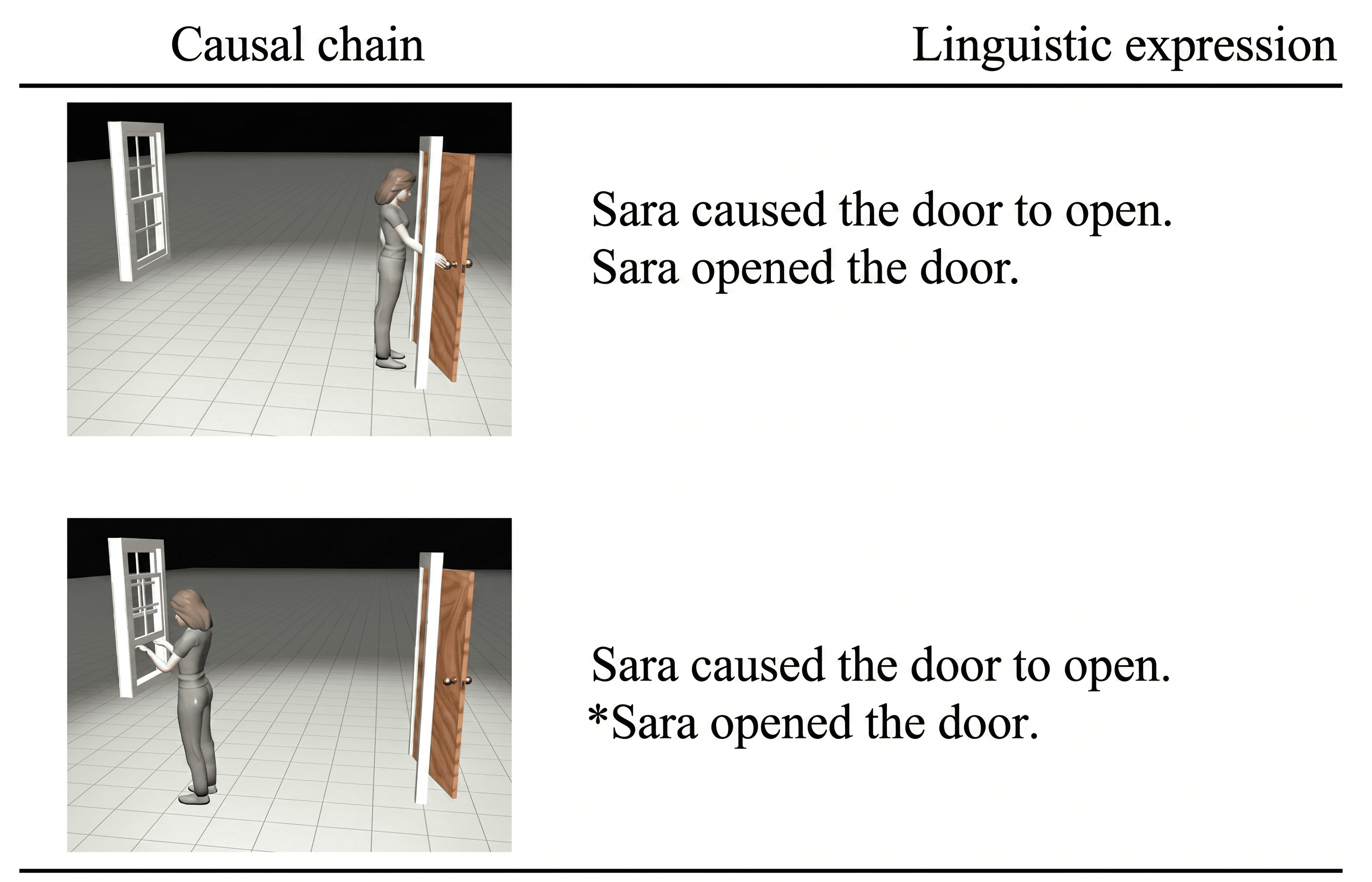}
    \caption{Linguistic descriptions of causal chains from \citet{Wolff2003}. Direct causation allows both forms; mediated causation requires periphrastic.}
    \label{fig:sara-door}
  \end{subfigure}
\end{figure}

Figure~\ref{fig:causal-chains} illustrates this: in the unmediated condition, a green ball directly contacts a red ball; in the mediated condition, an intermediate grey ball transmits the force.

Humans reliably map chain type to linguistic form: \textbf{lexical causatives} (``Sara opened the door'') are preferred for unmediated events, while \textbf{periphrastic causatives} (``Sara caused the door to open'') are preferred for mediated events. Figure~\ref{fig:sara-door} demonstrates this: when Sara directly pushes a door, both forms are acceptable; when she opens a window and a breeze causes the door to open, only the periphrastic form is felicitous.

\begin{figure}[h]
  \centering
  \includegraphics[width=0.65\linewidth]{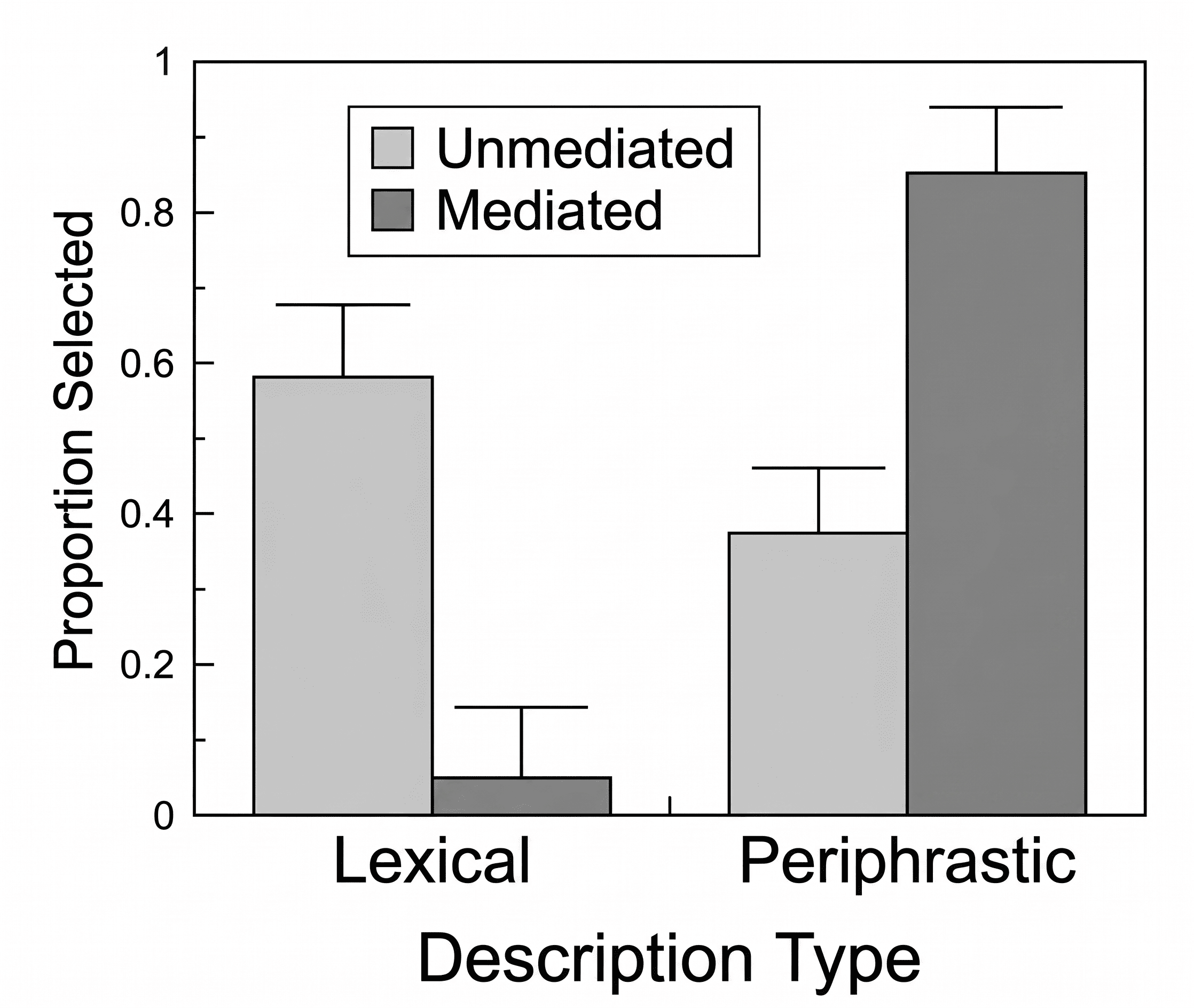}
  \caption{Human description preferences from \citet{Wolff2003}. For unmediated events, 
  participants chose lexical causatives ($\sim$58\%) over periphrastic ($\sim$38\%); 
  for mediated events, the pattern reversed sharply, with periphrastic descriptions 
  selected $\sim$85\% of the time and lexical only $\sim$5\%. This double dissociation 
  is the human baseline our NLI models fail to replicate.}
  \label{fig:wolff-results}
\end{figure}

Wolff's quantitative results (Figure~\ref{fig:wolff-results}) reveal a double dissociation: for unmediated events, participants selected lexical descriptions $\sim$58\% of the time; for mediated events, they selected periphrastic descriptions $\sim$85\% of the time. This asymmetry matches the M-Heuristic: when speakers choose the marked form over the unmarked alternative, listeners infer non-stereotypical causation.

We designed an NLI experiment to test whether transformers show the same causative-form sensitivity as humans, adapting Wolff's production paradigm to a comprehension task.

\subsubsection{Research Question and Hypotheses}

We test whether transformer-based NLI models exhibit sensitivity to the M-Heuristic:
\textbf{H$_1$}: Models employ the M-Heuristic, predicting Entailment when causative form matches manner (lexical+normal, periphrastic+unusual) and Contradiction when they mismatch.
\textbf{H$_0$}: Models treat lexical and periphrastic causatives as semantically equivalent, predicting predominantly Neutral.

\subsubsection{Linguistic Phenomena: Causative Alternation}

We focus on \textbf{ambitransitive verbs}, i.e., verbs with both an intransitive use (``The ice melted'') and a transitive use (``The chef melted the butter''), ensuring both lexical and periphrastic causatives are available. We selected 15 verbs across semantic domains: \textit{change of state} (melt, freeze, dry, dissolve, burn), \textit{breaking} (break, crack, shatter), \textit{motion} (move, roll, stop, start, sink), and \textit{aperture} (open, close). Premises were constructed as lexical (``The chef melted the butter'') or periphrastic (``The chef caused the butter to melt'') causatives.

\subsubsection{Methodology}

NLI requires determining whether a Premise entails, contradicts, or is neutral with respect to a Hypothesis \cite{Dagan2006}. Pragmatic entailment differs from logical entailment: ``Jo ate some of the cake'' pragmatically (but not logically) implies ``Jo didn't eat all of the cake.'' We test whether models treat manner hypotheses in the same spirit.

We paired each premise type with hypotheses describing either \textbf{normal/direct} or \textbf{unusual/indirect} manner, creating a 2$\times$2 design (Table~\ref{tab:conditions}). Expected labels derive from the M-Heuristic: unmarked lexical forms I-implicate stereotypical manner (Lexical+Normal $\Rightarrow$ Entailment; Lexical+Unusual $\Rightarrow$ Contradiction), while marked periphrastic forms M-implicate non-stereotypical manner (Periphrastic+Normal $\Rightarrow$ Contradiction; Periphrastic+Unusual $\Rightarrow$ Entailment).

\begin{table}[h]
\centering
\caption{Experimental Conditions and Expected Human Judgments}
\label{tab:conditions}
\begin{tabular}{clll}
\toprule
\textbf{Cond.} & \textbf{Premise Type} & \textbf{Hypothesis Type} & \textbf{Expected Label} \\
\midrule
1 & Lexical causative & Normal interpretation & \textbf{Entailment} \\
2 & Lexical causative & Unusual interpretation & \textbf{Contradiction} \\
3 & Periphrastic causative & Normal interpretation & \textbf{Contradiction} \\
4 & Periphrastic causative & Unusual interpretation & \textbf{Entailment} \\
\bottomrule
\end{tabular}
\end{table}

\subsubsection{Dataset Construction}

For each of the 15 verbs, we created 3 distinct contexts varying the agent and patient, tested across all 4 conditions, yielding 180 items. The dataset also includes 8 items instantiating the canonical Levinson examples directly (the Larry/car pair and an outlaw/sheriff pair, each tested across all 4 conditions), for a total of \textbf{188} test items in the released dataset. Table~\ref{tab:sample-data} presents sample items.

\begin{table}[h]
\centering
\caption{Sample Experimental Items}
\label{tab:sample-data}
\small
\begin{tabular}{p{2.8cm}p{3.2cm}p{3.8cm}l}
\toprule
\textbf{Premise Type} & \textbf{Premise} & \textbf{Hypothesis} & \textbf{Expected} \\
\midrule
Lexical & The chef melted the butter. & The chef did this in a normal, direct way. & Entailment \\
Periphrastic & The chef caused the butter to melt. & The chef did this in an unusual or indirect way. & Entailment \\
\bottomrule
\end{tabular}
\end{table}

The lexical form I-implicates stereotypical manner (e.g., ``Larry stopped the car'' $\Rightarrow$ by pressing the brake); the periphrastic form M-implicates non-standard means (e.g., ``Larry caused the car to stop'' $\Rightarrow$ by throwing an obstacle in its path).

We evaluated three transformer models fine-tuned on MNLI: \textbf{DeBERTa-v3-large} \cite{he2021deberta}, our primary model due to its disentangled attention mechanism; \textbf{RoBERTa-large} \cite{Liu2019} as an encoder-only baseline; and \textbf{BART-large} \cite{Lewis2020} for encoder-decoder comparison.

\subsection{Gemini Prompt Comparison}

To probe whether the failure of encoder NLI models reflects an \textit{absence} of pragmatic knowledge or an absence of \textit{spontaneous deployment} under default processing, we additionally evaluated \textbf{Gemini Flash-Lite} on the same causative items using three prompt regimes (full templates in Appendix~\ref{sec:gemini-prompts}). All prompts ask for a three-way judgment aligned with our pragmatic gold (Entailment vs.\ Contradiction vs.\ Neutral), instantiated as \texttt{ENTAILMENT} / \texttt{CONTRADICTION} / \texttt{NEUTRAL}.

\textbf{Direct NLI} presents a compact Natural Language Inference instruction: the model must decide whether the premise entails, contradicts, or is neutral with respect to the hypothesis, and answer with a single label word, mirroring standard NLI elicitation without additional scaffolding (Appendix~\ref{sec:gemini-prompts}).

\textbf{Chain-of-thought} retains the same underlying NLI task but instructs the model to reason about what the premise implies, including what a speaker would typically implicate by choosing the given phrasing, before committing to a label. The model is asked to output structured \texttt{REASONING:} and \texttt{LABEL:} fields so that intermediate steps can be inspected.

\textbf{Metalinguistic} supplies explicit background on the M-Heuristic (with canonical Larry/car examples contrasting lexical and periphrastic causatives), then asks for the same NLI judgment with \texttt{REASONING:} and \texttt{LABEL:} fields.

The motivating hypothesis: if accuracy rises sharply only when the M-Heuristic is stated metalinguistically, pragmatic knowledge is plausibly \textit{present} in the model weights yet not \textit{procedurally integrated} into the default decision procedure elicited by ordinary NLI prompts.

\section{Results}

Table~\ref{tab:overall-accuracy} and the subsections below compare model labels to pragmatic gold. No encoder NLI model produced M-Heuristic-consistent labels at scale. Table~\ref{tab:overall-accuracy} gives accuracy against pragmatic gold.

\begin{table}[h]
\centering
\caption{Overall Accuracy on M-Heuristic Inference Task}
\label{tab:overall-accuracy}
\begin{tabular}{lcc}
\toprule
\textbf{Model} & \textbf{Correct / Total} & \textbf{Accuracy (\%)} \\
\midrule
DeBERTa-v3-large & 0 / 188 & 0.0 \\
RoBERTa-large & 5 / 188 & 2.7 \\
BART-large & 0 / 188 & 0.0 \\
\bottomrule
\end{tabular}
\end{table}

\textbf{DeBERTa} scored 0\%, predicting ``Neutral'' on every pair regardless of form or hypothesis type, with no sensitivity to the lexical vs.\ periphrastic contrast. \textbf{RoBERTa} reached 2.7\% (5 Contradictions), far below both the 33\% three-class random baseline and the 50\% binary baseline applicable when gold labels are restricted to Entailment and Contradiction, so labels carry no reliable pragmatic signal. \textbf{BART} also scored 0\%, again skewed to Neutral.

Table~\ref{tab:cm-deberta} gives DeBERTa's confusion matrix: 188/188 Neutral. RoBERTa was 89.4\% Neutral; its five Contradiction hits are within chance noise. BART was 97.8\% Neutral. No model assigned Entailment to the two gold-Entailment conditions (Lexical+Normal, Periphrastic+Unusual).

\begin{table}[h]
\centering
\caption{Confusion Matrix: DeBERTa-v3-large}
\label{tab:cm-deberta}
\begin{tabular}{l|ccc}
\toprule
\textbf{Expected $\downarrow$ / Predicted $\rightarrow$} & \textbf{Entailment} & \textbf{Neutral} & \textbf{Contradiction} \\
\midrule
\textbf{Entailment} & 0 & 94 & 0 \\
\textbf{Contradiction} & 0 & 94 & 0 \\
\bottomrule
\end{tabular}
\end{table}

Gold labels omit Neutral entirely (form--manner pairing fixes Entailment vs.\ Contradiction). Model mass on Neutral (Table~\ref{tab:overall-accuracy}) therefore reflects no usable link between causative form and manner in their predictions. No model exceeded chance in any of the four conditions of Table~\ref{tab:conditions}.

\subsection{Gemini Prompt Comparison Results}

Table~\ref{tab:gemini-prompts} summarizes Gemini Flash-Lite accuracy against pragmatic gold on the same 188 items (labels parsed from model outputs).

\begin{table}[h]
\centering
\caption{Gemini Flash-Lite accuracy on pragmatic gold labels by prompt condition ($n = 188$)}
\label{tab:gemini-prompts}
\begin{tabular}{lcc}
\toprule
\textbf{Prompt condition} & \textbf{Correct / Total} & \textbf{Accuracy (\%)} \\
\midrule
Direct NLI & 0 / 188 & 0.0 \\
Chain-of-thought & 41 / 188 & 21.8 \\
Metalinguistic (M-Heuristic) & 188 / 188 & 100.0 \\
\bottomrule
\end{tabular}
\end{table}

Under \textbf{direct} prompting, the model predicted \texttt{NEUTRAL} for every item, matching DeBERTa's all-Neutral encoder behavior. Under \textbf{chain-of-thought}, many gold \texttt{Entailment} items are only partially recovered, while gold \texttt{Contradiction} items seldom receive \texttt{CONTRADICTION}; the model stays on \texttt{NEUTRAL} for that half of the design, so violation judgments stay sparse.

By contrast, the \textbf{metalinguistic} prompt yields perfect agreement with pragmatic gold. Manual inspection of traces shows item-specific reasoning across verb--patient contexts rather than a single template, which argues against pure label memorization.

\section{Discussion}

We find no evidence that state-of-the-art transformer models fine-tuned for NLI employ the M-Heuristic; we therefore fail to reject H$_0$. What follows interprets the findings, considers explanations, and places them in work on pragmatic competence in neural language models.

DeBERTa's all-Neutral run (Table~\ref{tab:overall-accuracy}) holds even with disentangled attention tuned to content and position. RoBERTa and BART show the same Neutral-heavy pattern. Models treat lexical and periphrastic causatives as truth-conditionally similar but miss pragmatic divergence.

The Gemini comparison refines this: direct prompts reproduce encoder-style all-Neutral behavior, so the gap is not encoder-only. Metalinguistic prompts reach 100\% with item-specific traces. Humans deploy M-inferences without such scaffolding; these models do not. To understand whether DeBERTa's failure is representational or a use failure, we ran probing experiments, detailed below.

\subsection{Step 1: Preliminary Positive Result}

We trained linear classifiers (logistic regression probes) at each of DeBERTa's 24 layers to distinguish lexical from periphrastic causatives based on the [CLS] token representation. The results, summarized in Table~\ref{tab:probing}, initially appeared encouraging.

\begin{table}[h]
\centering
\caption{Linear Probe Accuracy: Lexical vs. Periphrastic Causatives}
\label{tab:probing}
\begin{tabular}{lcl}
\toprule
\textbf{Layer Range} & \textbf{Accuracy} & \textbf{Initial Interpretation} \\
\midrule
Layer 0 (embedding) & 50.0\% & Chance level \\
Layers 1--3 & 80--99\% & Distinction emerges \\
\textbf{Layers 4--11} & \textbf{100.0\%} & \textbf{Perfect separation} \\
Layers 12--24 & 94--99\% & Slight degradation \\
\bottomrule
\end{tabular}
\end{table}

Probes achieved 100\% accuracy at layers 4--11 in distinguishing ``The chef melted the butter'' from ``The chef caused the butter to melt.'' This suggested a potential \textit{representation-use dissociation}: the model encodes the causative distinction but the NLI head fails to use it. Yet high probe accuracy alone does not show \textit{pragmatic} meaning, or it may track surface syntax alone.

\subsection{Step 2: Methodological Complication}
\label{sec:control}

To test whether probes detect causative pragmatics specifically or general syntactic complexity, we introduced \textbf{syntactic controls}: sentences with similar periphrastic structure but no M-implicature. We tested whether probes distinguish lexical causatives from constructions like ``The chef \textit{wanted/watched/allowed/expected} the butter to melt.'' If the original probe detects causative-specific information, it should perform at chance on these controls.

\begin{table}[h]
\centering
\caption{Probe Accuracy on Syntactic Controls by Layer}
\label{tab:controls}
\begin{tabular}{lcc}
\toprule
\textbf{Layer} & \textbf{Causative Probe} & \textbf{Control Probe} \\
\midrule
Layer 4 & 100.0\% & 87.5\% \\
Layer 8 & 100.0\% & 87.5\% \\
Layer 12 & 97.8\% & 87.5\% \\
Layer 16 & 96.7\% & 97.5\% \\
Layer 20 & 98.9\% & 97.5\% \\
\bottomrule
\end{tabular}
\end{table}

Results show probes achieve 87.5--97.5\% accuracy on syntactic controls, comparable to causative probes. This reveals the original 100\% accuracy was largely tracking \textbf{syntactic complexity} (simple vs. embedded clause structure), not causative pragmatics specifically. The apparent representation-use dissociation was an artifact of probe design.

One nuance: at layer 4, control accuracy (87.5\%) is lower than causative accuracy (100\%), with per-control-type accuracy near chance (55.6\%). Early layers may carry some causative-specific signal before the syntactic confound dominates. That does not restore the original interpretation.

\subsection{Step 3: Semantic Similarity}

We tested whether DeBERTa encodes pragmatic meaning. For 30 triplets, we compared the similarity between:
\begin{itemize}
    \item \textbf{Anchor}: A periphrastic causative (e.g., ``Larry caused the car to stop'')
    \item \textbf{Mediated description}: Indirect causation (``Larry threw an obstacle onto the road, forcing the car to halt'')
    \item \textbf{Unmediated description}: Direct causation (``Larry pressed the brake pedal firmly'')
\end{itemize}

If DeBERTa encodes the M-Heuristic's pragmatic content, anchors should cluster with mediated descriptions. Anchors cluster with \textbf{unmediated} descriptions in 29 of 30 cases ($p < 0.0001$, binomial test). That pattern runs \textit{opposite} to M-Heuristic predictions. DeBERTa's semantic space organizes causative sentences by literal event content rather than implied causal structure. The periphrastic anchor ``Larry caused the car to stop'' is closer to ``Larry pressed the brake pedal'' than to ``Larry threw an obstacle onto the road'', grouping by event type while ignoring the pragmatic signal entirely.

\begin{table}[h]
\centering
\caption{Semantic Similarity Results (Layer 12, n=30 triplets)}
\label{tab:similarity}
\begin{tabular}{lc}
\toprule
\textbf{Metric} & \textbf{Value} \\
\midrule
Anchor closer to mediated & 1/30 (3.3\%) \\
Anchor closer to unmediated & 29/30 (96.7\%) \\
Binomial test (vs. 50\%) & $p < 0.0001$ \\
Mean similarity (all pairs) & $\sim$0.999 \\
\bottomrule
\end{tabular}
\end{table}

\begin{figure}[h]
  \centering
  \includegraphics[width=\linewidth]{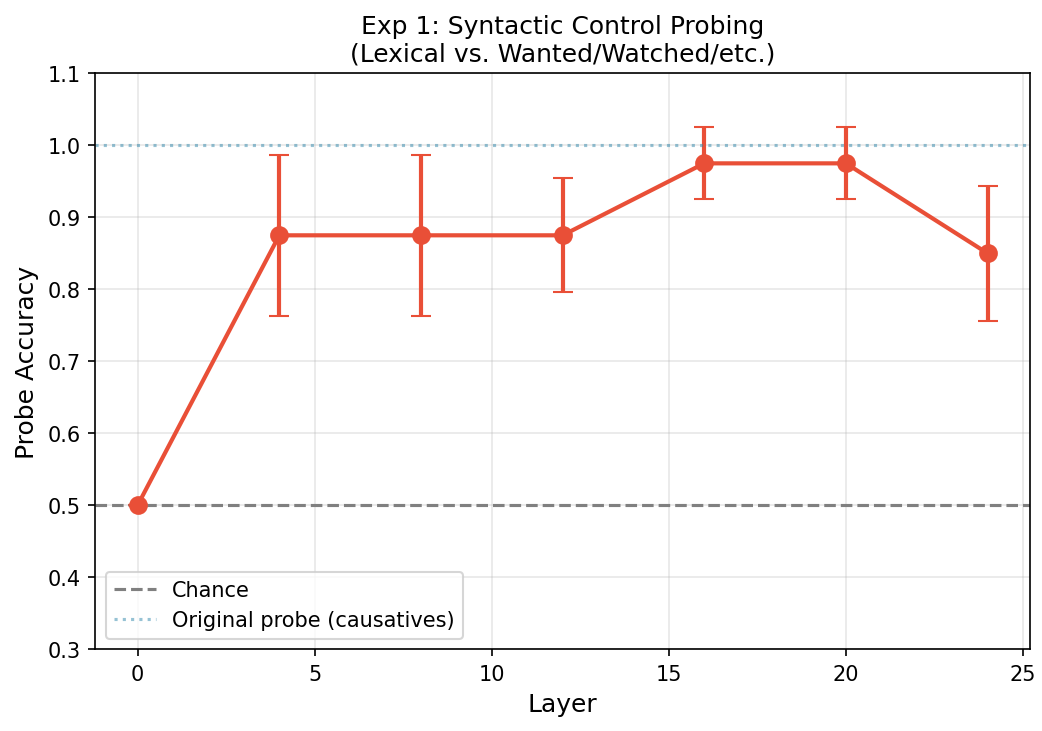}
  \caption{Probe accuracy on syntactic controls by layer. Probes achieve 87.5--97.5\% accuracy on non-causative periphrastic constructions, matching causative probe accuracy and indicating that the original 100\% result tracks syntactic complexity rather than causative pragmatics.}
  \label{fig:controls-probes}
\end{figure}

\begin{figure}[h]
  \centering
  \includegraphics[width=\linewidth]{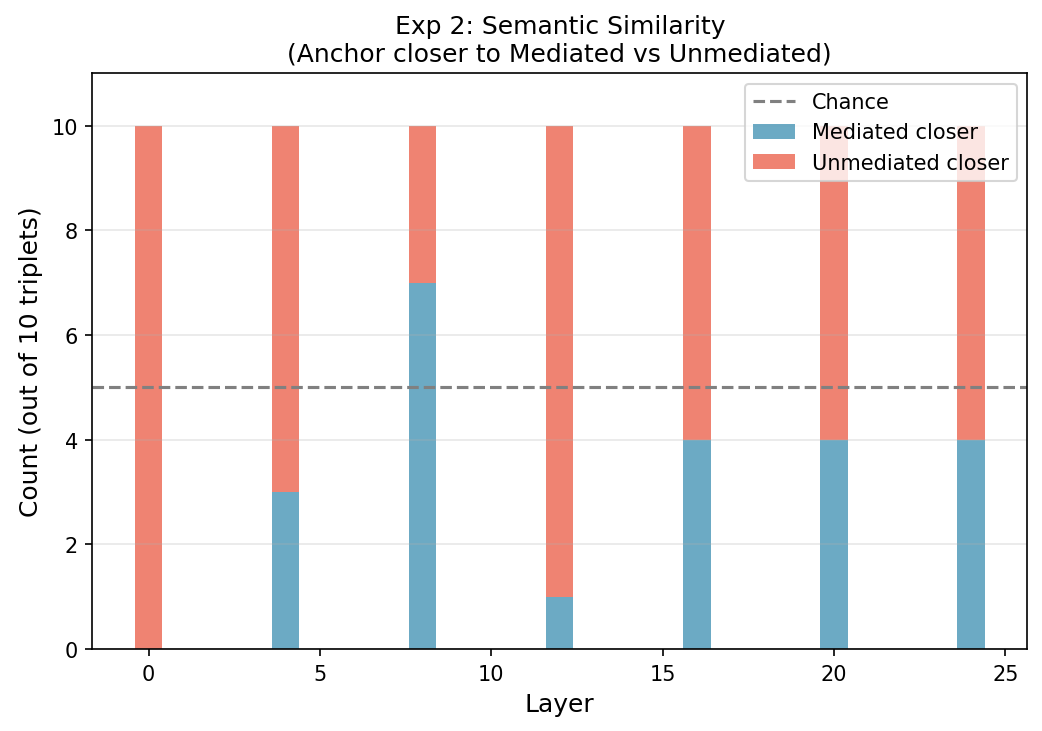}
  \caption{Semantic similarity results. Periphrastic causative anchors cluster with unmediated (direct) descriptions in 29/30 cases, showing that DeBERTa's representations encode the pragmatically \textit{wrong} grouping relative to M-Heuristic predictions.}
  \label{fig:controls-similarity}
\end{figure}

The high similarities ($\sim$0.999) across all pairs show sentence-level representations are so compressed that manner contrasts leave little measurable geometry. Figures~\ref{fig:controls-probes} and~\ref{fig:controls-similarity} visualize both control experiments.

The three experiments together read as a single story:

\begin{enumerate}
    \item \textbf{Behavioral level}: DeBERTa predicts Neutral uniformly, with no pragmatic inferences from form.
    \item \textbf{Representational level (probe)}: Initially appeared to encode the distinction.
    \item \textbf{Control probe}: Reveals this was syntactic complexity tracking, not causative-specific.
    \item \textbf{Semantic similarity}: Confirms representations organize around literal event content and, if anything, encode the \textit{opposite} of M-Heuristic implications.
\end{enumerate}

Beyond a simple representation--use split, \textbf{DeBERTa fails at both using pragmatic cues and encoding the M-Heuristic contrast in sentence representations}. Its layers track syntax and event semantics, not the form--meaning pairing that defines the M-Heuristic. Remediation would need more than a new classification head: training that instills marked-form--marked-meaning mappings at the representation level.

\subsection{Why Do Models Predict Neutral?}

Several factors likely contribute to the Neutral bias. First, NLI training corpora contain almost no M-implicature examples \cite{Jeretic2020}; models have never learned that linguistic form alone can determine inferential relationships. Second, M-implicatures require comparing the used form against an unused alternative (``Larry caused the car to stop'' implicates unusual manner only because the speaker \textit{could have said} ``Larry stopped the car''). Current architectures have no mechanism for representing such counterfactual alternatives. Third, the relationship between periphrastic structure and indirect causation derives from a general principle about markedness and communicative effort, not distributional co-occurrence statistics that transformers can learn.

We acknowledge alternative explanations for our results: (a) NLI may be poorly suited for eliciting pragmatic reasoning, since implicatures are defeasible and may not map cleanly onto the Entailment/Neutral/Contradiction trichotomy; (b) our ``gold'' labels derive from theory rather than human annotation, so reasonable annotators might disagree whether pragmatic inferences constitute entailments; (c) models may be behaving appropriately for a \textit{semantic} task, correctly recognizing that both causative forms are truth-conditionally compatible with any manner of action. Probing still shows representations grouped with the pragmatically \textit{wrong} descriptions, so the issue is deeper than task framing alone.

\section{Conclusion}

Transformer NLI models fine-tuned on MNLI do not behave as if they use the M-Heuristic on our 188-item causative test items: DeBERTa is all-Neutral, while RoBERTa and BART are Neutral-heavy (Table~\ref{tab:overall-accuracy}). Probes track syntax, not the targeted pragmatic contrast. Similarity space groups periphrastic anchors with unmediated descriptions, against M-Heuristic geometry. Gemini Flash-Lite reaches ceiling accuracy only with metalinguistic prompts and item-specific traces, consistent with knowledge that default NLI does not recruit. Closing the gap will require training that rewards spontaneous pragmatic inference, not merely exposure to the principles that govern it.

Future work might explore:
\begin{itemize}
    \item Expanding the probing dataset with additional verbs and contexts, using held-out verb classes for validation, to confirm that the observed representational separability is robust and not an artifact of the small sample size.
    \item Developing training data that explicitly annotates form-based implicatures.
    \item Evaluation on a broader range of M-implicature phenomena beyond causatives.
\end{itemize}

\section{Limitations}

\begin{enumerate}
    \item \textbf{Hypothesis phrasing}: Our hypotheses explicitly mention ``normal'' and ``unusual'' manner. Alternative phrasings (e.g., describing specific mechanisms) might yield different results, though we expect similar patterns.

    \item \textbf{Model selection}: We primarily evaluated three models fine-tuned on MNLI. Gemini Flash-Lite (Table~\ref{tab:gemini-prompts}) is \emph{prompt-dependent}: all-Neutral under direct NLI, partial chain-of-thought recovery, perfect scores when the principle is spelled out. That narrows how far one should generalize from surprisal-based ``struggle with M-implicatures'' claims \cite{Cong2024} to fixed ceilings on model knowledge. Systematic sweeps over families, scales, and instruction tuning are left to future work.

    \item \textbf{Zero-shot evaluation}: Our experiments used off-the-shelf NLI models without task-specific fine-tuning. Whether models \textit{could} learn M-Heuristic inference given appropriate training data remains an open question.

    \item \textbf{Metalinguistic prompt confound}: The metalinguistic condition explicitly states the M-Heuristic principle, so it remains difficult to fully distinguish pre-existing parametric knowledge from within-prompt adaptation. Transfer evaluation on non-causative M-implicature constructions would be needed to establish generality beyond the causative alternation.

\end{enumerate}

%%
%% Acknowledgments
\begin{acknowledgments}
  This research is supported by the project "Center of Excellence for Climate and Societal Change", project number PN-IV-P6-6.1-CoEx-2024-004. 
\end{acknowledgments}

\section{Ethics Statement}
This research did not involve human participants or any personally identifiable information. Consequently, no ethical approval was required, and the authors declare that there are no ethical issues or concerns associated with this work.

%% Declaration on Generative AI (required by CEUR-WS from January 2025)
\section*{Declaration on Generative AI}
The author(s) did not use any Generative AI tools in the preparation of this work. Therefore, no roles from the CEUR-WS GenAI Usage Taxonomy apply to the drafting or editing of this manuscript. 
However, as the subject of the research itself, Google Gemini Flash-Lite was accessed programmatically (via API) as an \emph{experimental system} under the fixed prompt templates in Appendix~\ref{sec:gemini-prompts} to obtain the NLI judgments analysed in Section~4. The author(s) confirm that the generated data has been reviewed and they take full responsibility for the published content.

%%
%% Bibliography
\bibliography{sample-ceur}

%%
%% Appendix: Gemini prompts
\appendix
\section{Gemini prompt templates}
\label{sec:gemini-prompts}

\noindent
In all conditions, the premise and hypothesis strings from the dataset were inserted as two lines immediately after the fixed instructions (quoted in the same format as in the experiments). The three regimes below differ only in the scaffolding shown here.

\subsection{Direct NLI}
\begin{lstlisting}[basicstyle=\ttfamily\footnotesize,breaklines=true,frame=single]
You are performing a Natural Language Inference task.
Given a Premise and a Hypothesis, determine whether the
Premise ENTAILS, CONTRADICTS, or is NEUTRAL with respect
to the Hypothesis.

Premise: "<premise>"
Hypothesis: "<hypothesis>"

Answer with exactly one word: ENTAILMENT, CONTRADICTION, or NEUTRAL.
\end{lstlisting}

\subsection{Chain-of-thought}
\begin{lstlisting}[basicstyle=\ttfamily\footnotesize,breaklines=true,frame=single]
You are performing a Natural Language Inference task.

Given a Premise and a Hypothesis, reason carefully about
what the Premise implies before deciding whether it
ENTAILS, CONTRADICTS, or is NEUTRAL with respect to
the Hypothesis.

Consider not just what the sentence literally says, but
what a speaker would typically imply by choosing that
particular phrasing.

Premise: "<premise>"
Hypothesis: "<hypothesis>"

Think step by step, then give your final answer as:
REASONING: [your reasoning]
LABEL: [ENTAILMENT / CONTRADICTION / NEUTRAL]
\end{lstlisting}

\subsection{Metalinguistic (M-Heuristic)}
\begin{lstlisting}[basicstyle=\ttfamily\footnotesize,breaklines=true,frame=single]
In linguistics, the M-Heuristic states that marked
(structurally complex) linguistic forms implicate marked
(unusual, indirect) meanings. For example:

- "Larry stopped the car" (lexical causative) implies
  direct, stereotypical causation (e.g., pressing the brake).
- "Larry caused the car to stop" (periphrastic causative)
  implies indirect or unusual causation (e.g., throwing
  an obstacle in the path.

Given this principle, determine whether the following
Premise ENTAILS, CONTRADICTS, or is NEUTRAL with respect
to the Hypothesis.

Premise: "<premise>"
Hypothesis: "<hypothesis>"

REASONING: [your reasoning]
LABEL: [ENTAILMENT / CONTRADICTION / NEUTRAL]
\end{lstlisting}

\end{document}